\documentclass[11pt, a4paper]{article}

\usepackage[absolute,overlay]{textpos}

\usepackage{hyperref}

\hypersetup{
    colorlinks=true,
    linkcolor=blue,
    filecolor=magenta,      
    urlcolor=cyan,
    pdftitle={Your Paper Title},
    pdfpagemode=FullScreen,
}

\usepackage[
    a4paper,
    top=2.5cm, 
    bottom=2.5cm, 
    left=3cm, 
    right=3cm, 
    footskip=1.2cm
]{geometry}

\usepackage{microtype}     
\usepackage{booktabs}      
\usepackage{titlesec}      
\usepackage{hyperref}      
\usepackage{xcolor}        

\usepackage[numbers]{natbib} 
\usepackage[T1]{fontenc}    
\usepackage{hyperref}       
\usepackage{url}            
\usepackage{booktabs}       
\usepackage{amsfonts}       
\usepackage{nicefrac}       
\usepackage{microtype}      
\usepackage{xcolor}         

\usepackage{times}
\usepackage{latexsym}
\usepackage{tabularx}
\usepackage{amsmath}
\usepackage{amssymb}
\usepackage{enumitem}
\usepackage{graphicx}
\usepackage{subcaption}
\usepackage{pifont}
\usepackage{xcolor}
\usepackage{titletoc}
\usepackage{colortbl}
\usepackage{wrapfig}  
\usepackage{lipsum} 
\usepackage{multirow}
\usepackage{listings}
\usepackage{array}
\usepackage{booktabs} 
\usepackage{caption}  
\usepackage{multicol}
\usepackage{xurl}
\usepackage{inconsolata}
\usepackage{ulem}
\usepackage{array}
\usepackage[ruled,linesnumbered,noend]{algorithm2e}
\usepackage{algpseudocode}
\usepackage{float}
\usepackage[most]{tcolorbox} 
\usepackage{authblk}
\usepackage{forest}

\makeatletter
\newcommand{\nofootnote}[1]{%
    \begingroup
    \renewcommand\@makefntext[1]{\parindent 1em\noindent##1}
    \footnotetext{#1}
    \endgroup
}
\makeatother

\usepackage{lipsum}
\usepackage[page]{appendix}

\renewcommand{\appendixtocname}{List of appendices}

\makeatletter
\let\oldappendix\appendices

\renewcommand{\appendices}{%
  \clearpage
  \renewcommand{\thesection}{\Roman{section}}
  \let\tf@toc\tf@app
  \addtocontents{app}{\protect\setcounter{tocdepth}{2}}
  \immediate\write\@auxout{%
    \string\let\string\tf@toc\string\tf@app^^J
  }
  \oldappendix
}%

\newcommand{\listofappendices}{%
  \begingroup
  \renewcommand{\contentsname}{\appendixtocname}
  \let\@oldstarttoc\@starttoc
  \def\@starttoc##1{\@oldstarttoc{app}}
  \tableofcontents
  \endgroup
}

\makeatother

\definecolor{myblue}{RGB}{32,116,184}
\definecolor{mybgblue}{RGB}{33,114,180}
\definecolor{saki}{HTML}{7799CC}
\definecolor{amber}{rgb}{1.0, 0.75, 0.0}

\hypersetup{
    colorlinks=true,
    linkcolor=red,
    citecolor=blue,
    filecolor=magenta,      
    urlcolor=blue,
linktocpage}

\title{PRBench: End-to-end Paper Reproduction in Physics Research}

\author[1]{\textbf{Shi~Qiu}}
\author[]{\textbf{Junyi~Deng}}
\author[]{\textbf{Yiwei~Deng}}
\author[]{\textbf{Haoran~Dong}}
\author[]{\textbf{Jieyu~Fu}}
\author[]{\textbf{Mao~Li}}
\author[]{\textbf{Zeyu~Li}}
\author[]{\textbf{Zhaolong~Zhang}}
\author[]{\textbf{Huiwen~Zheng}}
\author[]{\textbf{Leidong~Bao}}
\author[]{\textbf{Anqi~Lv}}
\author[]{\textbf{Zihan~Mo}}
\author[]{\textbf{Yadi~Niu}}
\author[]{\textbf{Yiyang~Peng}}
\author[]{\textbf{Yu~Tian}}
\author[]{\textbf{Yili Wang}}
\author[]{\textbf{Ziyu Wang}}
\author[]{\textbf{Zi-Yu~Wang}}
\author[]{\textbf{Jiashen~Wei}}
\author[]{\textbf{Liuheng~Wu}}
\author[]{\textbf{Aoran~Xue}}
\author[]{\textbf{Leyi~Yang}}
\author[]{\textbf{Guanglu~Yuan}}
\author[]{\textbf{Xiarui~Zhan}}
\author[]{\textbf{Jingjun~Zhang}}
\author[]{\textbf{Zifan~Zheng}}
\author[]{\textbf{Pengfei~Liu}}
\author[]{\textbf{Linrui~Zhen}}
\author[]{\textbf{Kaiyang~Li}}
\author[]{\textbf{Qichang~Li}}
\author[]{\textbf{Ziheng~Zhou}}
\author[]{\textbf{Guo-En~Nian}}
\author[]{\textbf{Yunwei~Xiao}}

\author[]{\textbf{Qing-Hong Cao}}
\author[]{\textbf{Linjie Dai}}
\author[]{\textbf{Xu Feng}}
\author[]{\textbf{Peng Gao}}
\author[]{\textbf{Ying Gu}}
\author[]{\textbf{Chang Liu}}
\author[]{\textbf{Jia Liu}}
\author[2]{\textbf{Ming-xing Luo}}
\author[]{\textbf{Yan-Qing Ma}}
\author[]{\textbf{Liang-You Peng}}
\author[]{\textbf{Huichao Song}}
\author[]{\textbf{Shufeng Wang}}
\author[]{\textbf{Chenxu Wang}}
\author[]{\textbf{Tao Wang}}
\author[]{\textbf{Yi-Nan Wang}}
\author[]{\textbf{Chengyin Wu}}
\author[]{\textbf{Pengwei Zhao}}
\author[1,†]{\textbf{Hua Xing Zhu}}

\affil[1]{School of Physics, Peking University, China}
\affil[2]{Beijing Computational Science Research Center, China} 

\vspace{-2em}

\begin{document}

\maketitle
\begin{textblock*}{5cm}(14cm,1.5cm) 
    \raggedleft \small RISE-AGI-2026-002
\end{textblock*}

\nofootnote{†Contact: zhuhx@pku.edu.cn\\
Project homepage: \url{https://prbench.phybench.cn/}}

\begin{abstract}
AI agents powered by large language models exhibit strong reasoning and problem-solving capabilities, enabling them to assist scientific research tasks such as formula derivation and code generation. 
However, whether these agents can reliably perform end-to-end reproduction from real scientific papers remains an open question.
We introduce PRBench, a benchmark of 30 expert-curated tasks spanning 11 subfields of physics. 
Each task requires an agent to comprehend the methodology of a published paper, implement the corresponding algorithms from scratch, and produce quantitative results matching the original publication.
Agents are provided only with the task instruction and paper content, and operate in a sandboxed execution environment.
All tasks are contributed by domain experts from over 20 research groups at the School of Physics, Peking University, each grounded in a real published paper and validated through end-to-end reproduction with verified ground-truth results and detailed scoring rubrics.
Using an agentified assessment pipeline, we evaluate a set of coding agents on PRBench and analyze their capabilities across key dimensions of scientific reasoning and execution.
The best-performing agent, OpenAI Codex powered by \texttt{GPT-5.3-Codex}, achieves a mean overall score of 34\%.
All agents exhibit a zero end-to-end callback success rate, with particularly poor performance in data accuracy and code correctness.
We further identify systematic failure modes, including errors in formula implementation, inability to debug numerical simulations, and fabrication of output data.
Overall, PRBench provides a rigorous benchmark for evaluating progress toward autonomous scientific research.
\end{abstract}
\section{Introduction}
\label{sec:intro}

\begin{figure}[h]
    \centering
    \includegraphics[width=\linewidth]{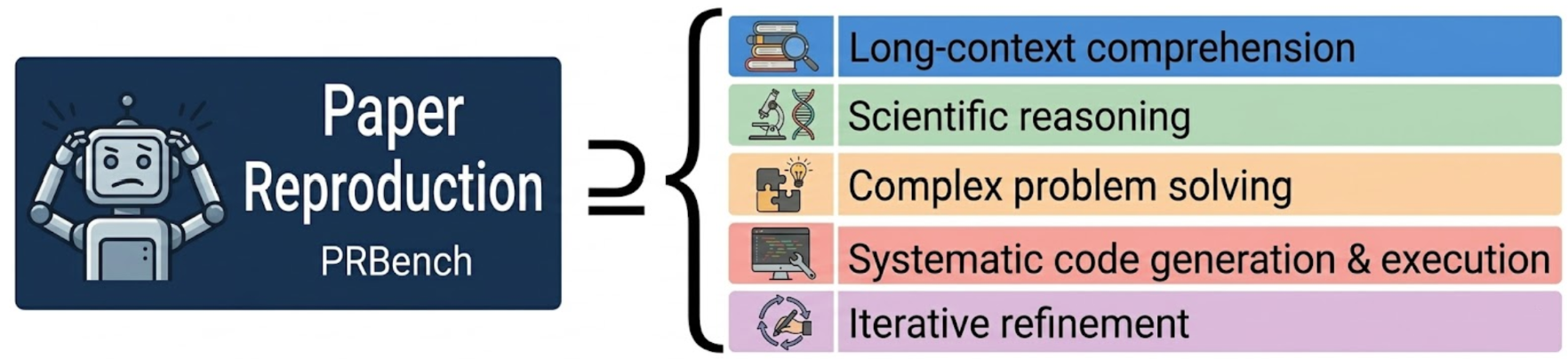}
    \caption{Overview of paper reproduction. As stated, reproducing computational results from a published paper is a comprehensive and demanding task.}
    \label{fig:placeholder}
\end{figure}
Recent advances in large language models (LLMs) have enabled AI agents with strong reasoning and systematic problem-solving capabilities, making them increasingly useful for assisting scientific research. 
Agents can now derive mathematical formulas~\citep{trinh2024alphageometry, he2024olympiadbench}, generate and debug scientific code~\citep{chen2021codex, roziere2023code}, propose experimental designs~\citep{boiko2023autonomous}, and support discovery across scientific domains~\citep{romera2024mathematical,novikov2025alphaevolve}.

However, it remains unclear whether AI agents can reliably perform end-to-end reproduction starting from a scientific paper alone. 
In physics, reproducing computational results from a published paper is a comprehensive and demanding task. 
It requires the agent to extract the underlying methodology from the original paper, implement the corresponding algorithms from scratch, and execute the full pipeline to obtain results consistent with the original work. 
Such a process demands the coordinated integration of multiple capabilities, including long-context comprehension, scientific reasoning, complex problem solving, systematic code generation and execution, and iterative refinement.

Existing benchmarks capture only partial aspects of this process. 
Prior work evaluates isolated capabilities such as code generation, bug fixing, or scientific reasoning~\citep{wang2026frontierscience,chen2024scienceagentbench,jimenez2024swebench,tian2024scicode,qiu2025phybenchholisticevaluationphysical}, but does not assess whether agents can carry out the full end-to-end workflow. 
Moreover, these benchmarks provide limited support for diagnosing failure modes across different stages of the reproduction process. 
As a result, current evaluations fail to distinguish between agents that merely \textit{interpret} a paper and those that can \textit{faithfully execute} it to obtain verifiable results.

We introduce \textbf{PRBench} (\textbf{P}aper \textbf{R}eproduction \textbf{Bench}mark) to address these limitations.
PRBench consists of 30 expert-curated tasks derived from published physics papers spanning 11 subfields, including lattice gauge theory, quantum optics, nuclear physics, plasma physics, and condensed matter physics.
All tasks are sourced from over 20 research groups at the School of Physics, Peking University.
Each task is manually validated by domain experts, who perform end-to-end reproduction of the original results and provide comprehensive metadata, including core methodology, reference implementations, verified ground-truth results, and detailed scoring rubrics.

Our evaluation framework follows the \textbf{Agentified Agent Assessment (AAA)} paradigm~\citep{agentbeats}, and is implemented within a sandboxed execution environment.
Using an automated grading agent with human-provided metadata, we evaluate agent performance across four dimensions: methodology understanding, code implementation correctness, data reproduction accuracy, and task completion.

We evaluate a diverse set of AI agents on PRBench. 
The best-performing agent, OpenAI Codex powered by \texttt{GPT-5.3-Codex}, achieves an overall score of 34\%. 
Most notably, the end-to-end callback rate remains \textbf{zero}, indicating that none of the evaluated agents can reliably reproduce correct results from a given paper. 
We further identify several systematic failure modes, including incorrect formula implementation, inability to debug numerical simulations, and fabrication of output data to satisfy output-format requirements.

Our contributions are as follows:
\begin{itemize}[leftmargin=*,itemsep=2pt]
\item \textbf{A high-quality, expert-validated benchmark.}
PRBench consists of end-to-end paper reproduction tasks sourced from real research projects.
All tasks are validated by domain experts, who perform rigorous reproduction and provide comprehensive metadata, including core methodology, reference implementations, verified ground-truth results, and detailed scoring rubrics.

\item \textbf{An agentified evaluation framework.}
We introduce a fully agentified evaluation pipeline within a sandboxed execution environment, where agents are required to autonomously complete the full workflow from paper understanding to code generation and numerical result generation.
This design ensures secure and controlled execution, enabling rigorous, reliable, and scalable evaluation of end-to-end scientific workflows.

\item \textbf{A comprehensive analytical taxonomy.}
We propose a unified taxonomy for both evaluation and failure analysis. 
On the evaluation side, it decomposes agent performance into methodology understanding, code correctness, data reproduction accuracy, and task completion. 
On the analysis side, it categorizes failure modes based on agent execution behavior, including data fabrication and errors in translating methodology into correct implementations.

\end{itemize}
\section{Related Work}
\label{sec:related}

\paragraph{Scientific AI and LLMs for Science.}
AI has made significant strides in scientific domains.
AlphaFold~\citep{jumper2021alphafold} revolutionized protein structure prediction, while specialized models have advanced materials science~\citep{merchant2023scaling}, weather forecasting~\citep{lam2023graphcast}, and mathematical reasoning~\citep{trinh2024alphageometry}.
In the LLM space, GPT-4 has been shown to assist with scientific workflows~\citep{ai4science2023impact}, and autonomous agents like Coscientist~\citep{boiko2023autonomous} can plan and execute simple chemistry experiments.
However, these systems typically operate within constrained domains with specialized training data, rather than attempting general-purpose reproduction of diverse research papers.

\paragraph{Benchmarks for Scientific Reasoning.}
Several benchmarks evaluate scientific reasoning capabilities of LLMs.
SciCode~\citep{tian2024scicode} tests the ability to generate code for scientific computing tasks drawn from research papers, but focuses on individual computational subroutines rather than full paper reproduction.
ScienceAgentBench~\citep{chen2024scienceagentbench} evaluates agents on data-driven scientific discovery tasks.
GPQA~\citep{rein2024gpqa} provides graduate-level science questions requiring deep domain knowledge.
PhyBench~\citep{qiu2025phybenchholisticevaluationphysical} focuses on physical intuition and formula derivation.
OlympiadBench~\citep{he2024olympiadbench} evaluates mathematical and physics problem-solving.
FrontierScience~\citep{wang2026frontierscience} extends this landscape with expert-level scientific tasks designed to probe frontier research capabilities.
These benchmarks test important aspects of scientific competence but none captures the full pipeline of reading a paper, implementing its methods, and reproducing its quantitative results.

\paragraph{Agentified Assessment for Complex Tasks.}
Most existing benchmarks rely on static evaluation protocols, such as exact matching, rule-based scoring, or model-judge evaluation~\citep{chen2021codex,he2024olympiadbench,qiu2025phybenchholisticevaluationphysical,chen2024scienceagentbench}. However, these approaches are difficult and costly for complex, agent-based evaluation where integrated environment and diverse output is considered.
Therefore, recent work has begun to explore agentified evaluation frameworks, where multiple agents are used to coordinate task execution and assessment.
In particular, the Agentified Agent Assessment (AAA) paradigm~\citep{agentbeats} based on Agent-to-agent(A2A)~\citep{google2025a2a} protocol introduces a structured approach in which a grading agent interacts with a task-solving agent to perform dynamic, context-aware evaluation.
Such designs are especially beneficial for complex, long-horizon tasks, as they enable flexible assessment beyond static metrics and allow evaluation to incorporate intermediate reasoning, execution traces, and structured feedback.
Thus, PRBench builds on the agentified assessment paradigm to enable rigorous evaluation under end-to-end scientific reproduction scenarios, where correctness depends not only on final outputs but also on faithful implementation, execution behavior, and adherence to underlying scientific methodology.

\section{The PRBench Benchmark}
\label{sec:benchmark}

\subsection{Overview}

PRBench is designed to evaluate AI agents on the end-to-end reproduction of computational results from scientific papers in physics.
The benchmark focuses on papers where the main results rely on non-trivial computational modeling or numerical simulation, rather than purely analytical derivations.
Each task requires an agent to read a real scientific paper, understand the underlying methodology, implement the described algorithms, execute the computation, and generate quantitative outputs that reproduce the results reported in the original publication.

Overall, PRBench contains 30 tasks spanning 11 subfields of physics, including quantum chromodynamics (QCD), quantum optics, nuclear physics, plasma physics, and condensed matter physics, as detailed in Table~\ref{tab:task_taxonomy}.

\begin{table}[htbp]
\centering
\caption{Distribution of PRBench tasks across physics subfields.
The benchmark contains 30 tasks spanning 11 subfields of physics.}
\label{tab:task_taxonomy}
\small
\begin{tabularx}{\linewidth}{Xc|Xc}
\toprule
\textbf{Subfield} & \textbf{Count} & \textbf{Subfield} & \textbf{Count} \\
\midrule
Quantum Optics      & 4 & Strong-Field Physics              & 2 \\
Lattice Gauge Theory / QCD                  & 3 & Condensed Matter / Many-Body      & 2 \\
Nuclear Physics                 & 3 & High-Energy Phenomenology         & 2 \\
Plasma Physics                  & 3 & General Relativity / Astrophysics & 2 \\
Quantum Computing / Ion Traps   & 3 & Atomic / Molecular Physics        & 2 \\
Mathematical Physics            & 1 & Computational Electrodynamics     & 2 \\
Heavy-Ion Physics               & 1 &                                   &   \\
\bottomrule
\end{tabularx}
\end{table}

All tasks are contributed by research groups affiliated with the School of Physics at Peking University, representing more than 20 active research groups.
Each task is curated and validated by domain experts who ensure that the underlying research problems are scientifically meaningful, computationally reproducible, and representative of real frontier research workflows.

\subsection{Task Curation Process}
\label{sec:curation}

The task curation follows a multi-stage process, as illustrated in Figure~\ref{fig:curation_pipeline}, to ensure both scientific validity and evaluation rigor:

\begin{figure}
    \centering
    \includegraphics[width=\linewidth]{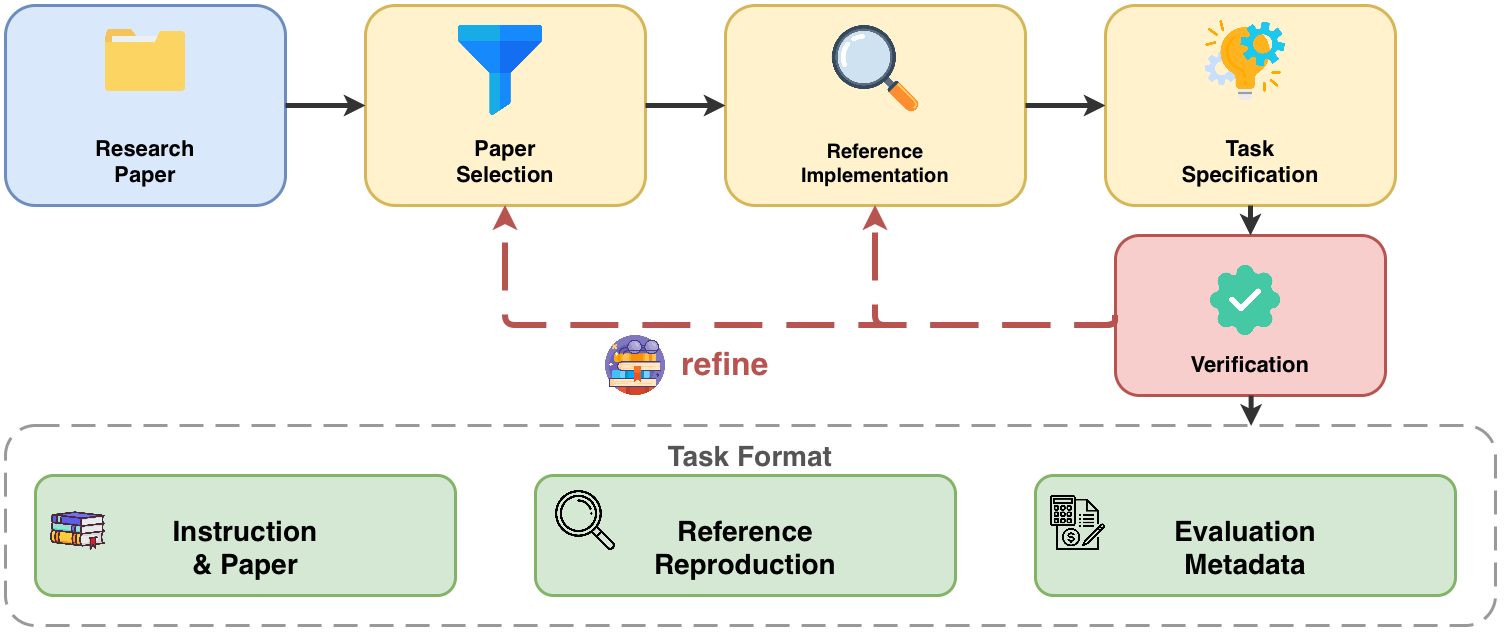}
    \caption{Overview of the PRBench task curation pipeline. The process consists of four stages: paper selection, expert reference reproduction, task specification, and independent verification, ensuring that each task is executable, scientifically grounded, and suitable for rigorous evaluation.}
    \label{fig:curation_pipeline}
\end{figure}

\begin{enumerate}[leftmargin=*,itemsep=2pt]

\item \textbf{Paper Selection}.
Research groups nominate candidate papers through internal discussion.
Selected papers must contain reproducible and scientifically meaningful computational results, supported by a sufficient number of figures or tables that serve as evaluation targets.
We focus on problems involving non-trivial numerical computation, such as simulations, parameter sweeps, or data-driven analysis, rather than purely analytical derivations.

To ensure reliable reproduction, the selected papers must provide a sufficiently detailed and self-contained description of the computational methodology, without relying heavily on external references for key implementation steps.
All tasks are additionally screened for computational feasibility, ensuring that they can be executed within a few hours in a sandboxed execution environment.
Further details are provided in Appendix~\ref{app:paper_selection}.

\item \textbf{Reference Implementation}.
For each selected paper, domain experts perform end-to-end reproduction and develop a reference implementation, including executable code and corresponding numerical outputs.
These implementations reproduce the key figures and tables from the original publication and serve as the ground-truth reference for evaluation.
While ensuring correctness, the reference outputs may include higher-resolution data to support more precise comparison.

\item \textbf{Task Specification}.
Each task is formalized into a structured specification (Section~\ref{sec:task_format}).
Outputs from the reference implementation are converted into standardized CSV files (Appendix~\ref{app:csv-reason}), enabling quantitative comparison between agent-generated results and ground truth.

The task specification includes the agent-visible instruction and a set of evaluation metadata, including methodological descriptions, expected outputs, and scoring criteria.
This metadata encodes both the numerical targets and the underlying physical and methodological constraints, allowing the evaluation to assess not only correctness of results but also consistency with the intended scientific procedure.

\item \textbf{Verification}.
Each task is independently verified by a domain expert.
The verifier checks that the reproduced outputs are consistent with the original publication and conform to the expected physical behavior.
They also validate that the extracted methodology and reference implementation faithfully reflect the procedures described in the paper.

During this stage, the evaluation metadata and scoring criteria are refined to ensure that the assessment captures methodological correctness, numerical accuracy, and physical plausibility.

\end{enumerate}

\subsection{Task Format}
\label{sec:task_format}

Each task in PRBench is defined by a set of expert-generated metadata that together specify both the task setup and the evaluation procedure.
This metadata consists of the following components:

\begin{itemize}[leftmargin=*,itemsep=2pt]

\item \textbf{Task Instruction and Source Paper.}
This component contains the task description with the full content of the referenced research paper.
The instruction specifies the target outputs, required formats, input parameters, and any constraints on the computational environment.
The source paper is provided as the only information accessible to the task-solving agent.

\item \textbf{Reference Implementation.}
Domain experts perform end-to-end reproduction of each task and provide a reference implementation, including executable code and generated outputs.
This component represents the human-validated reproduction of the original work and is used by the grading agent as the ground-truth reference for evaluation.

\item \textbf{Detailed Scoring Rubric.}
The scoring rubric specifies fine-grained evaluation criteria, including methodological checkpoints, expected numerical outputs, and weighting of different aspects of the implementation.
This design assigns higher importance to critical implementation details, improving the physical reliability and domain-specific rigor of the evaluation.

\end{itemize}

All components are provided, with strict separation between agent-visible inputs and evaluation resources.
This ensures that agents must interpret and implement the scientific methodology, rather than relying on access to ground-truth solutions. The task format design thus aligns well with the agentified evaluation framework. Additional details of the task format are provided in Appendix~\ref{app:task_format}.

\subsection{Evaluation Framework}
\label{sec:pipeline}

\begin{figure}[t]
    \centering
    \includegraphics[width=0.8\linewidth]{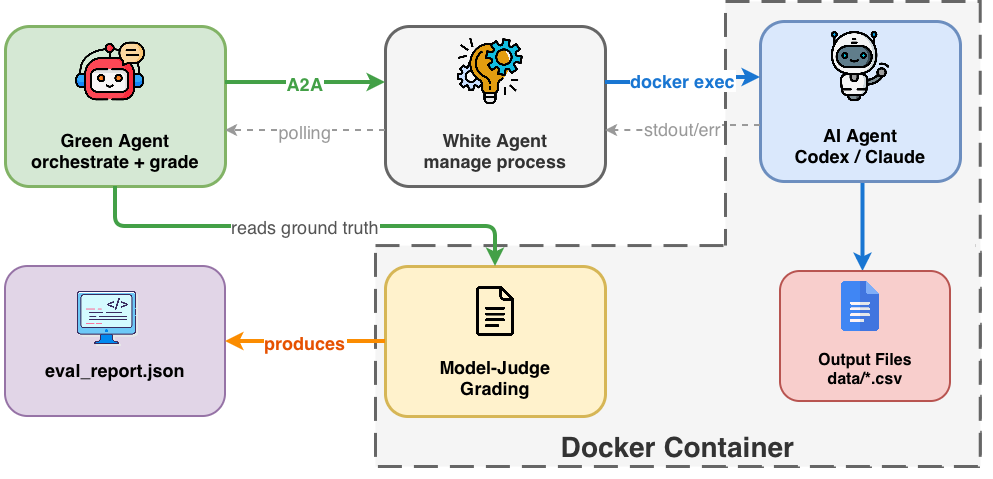}
    \caption{Overview of the PRBench evaluation pipeline. A green agent orchestrates the process and performs grading, while a white agent executes the task inside a sandboxed Docker environment.}
    \label{fig:eval-pipeline}
\end{figure}

PRBench is evaluated using an \textbf{agentified assessment framework} based on the Agent-to-Agent (A2A) communication protocol and the Agentified Agent Assessment (AAA) paradigm~\citep{agentbeats}.
The framework employs two coordinated agents: a \textit{white agent} responsible for task solving and execution, and a \textit{green agent} responsible for orchestration and evaluation, as detailed in Figure~\ref{fig:eval-pipeline}

For each task, the white agent receives the task instruction together with the full paper content, analyzes the methodology, generates the required code, and executes the computation inside a sandboxed execution environment implemented via Docker.
The green agent manages the evaluation process, dispatching instructions to the white agent, monitoring execution through periodic polling, and triggering evaluation once the task is completed.

All executions are performed within sandboxed environments with strict isolation, ensuring reproducibility and preventing information leakage.
After execution, the green agent invokes grading within the same environment, comparing the generated outputs against ground-truth metadata provided by domain experts.

The containerized architecture ensures strict isolation between task execution and evaluation, guaranteeing fairness and consistency of the assessment.
In addition, the framework supports parallel execution across tasks through independent container instantiation, enabling scalable and efficient benchmarking. Additional implementation details are provided in Appendix~\ref{app:pipeline}.

\section{Experiments}
\label{sec:experiment}

\subsection{Experimental Setup}

\paragraph{Agents Under Evaluation.}
We evaluate several task-solving agents based on different frontier models and execution frameworks.
The evaluated configurations include OpenAI Codex powered by \texttt{GPT-5.3-Codex}, OpenCode powered by \texttt{GPT-5.3-Codex}, and OpenCode-based agents powered by \texttt{GLM-5}, \texttt{Kimi K2.5}, \texttt{DeepSeek V3.2}, and \texttt{Minimax 2.7}.

For each task, the agent receives the task instruction with the full paper content, analyzes the methodology, generates the required implementation, executes the computation, and produces the final numerical outputs.
To reduce randomness in agent behavior, each task is executed three independent times for every agent configuration, and the reported scores are averaged across runs.

\paragraph{Evaluation Dimensions}
\label{sec:grading}

Each task is evaluated across four dimensions that together measure the agent's ability to reproduce scientific results:

\begin{enumerate}[leftmargin=*,itemsep=2pt]

\item \textbf{Methodology Understanding} \textit{(weight: 0.05).}  
Whether the agent correctly identifies the key formulas, algorithms, and physical observables described in the paper.

\item \textbf{Code Implementation Correctness} \textit{(weight: 0.30).}  
Whether the generated implementation faithfully realizes the computational procedure described in the paper, including algorithmic structure and numerical methods. 
Evaluation is guided by expert-provided scoring rubrics, which emphasize critical implementation details (e.g., correct formulation of key steps, numerical routines, and structural design), rather than superficial code similarity, thereby avoiding over-reliance on purely syntactic or stylistic differences.

\item \textbf{Data Reproduction Accuracy} \textit{(weight: 0.60).}  
How closely the generated numerical outputs match the reference data derived from the original publication.
Since numerical precision and sampling resolution may vary across implementations, evaluation considers not only pointwise agreement but also consistency with expected physical behavior, using task-specific criteria that account for acceptable deviations in scale, trend, and tolerance.

\item \textbf{Task Completeness} \textit{(weight: 0.05). } 
Whether all required artifacts (analysis, implementation, and output data) are produced and non-trivial.

\end{enumerate}

The overall score is computed as the weighted sum:

\begin{equation}
S_{\text{overall}} =
0.05\,S_{\text{method}} +
0.30\,S_{\text{code}} +
0.60\,S_{\text{data}} +
0.05\,S_{\text{complete}} .
\end{equation}

\paragraph{End-to-End Callback Rate.}
Beyond averaged scores, we introduce the \textbf{End-to-End Callback Rate} to measure whether an agent truly completes the reproduction task.
A run is considered successful if all evaluation dimensions achieve a score greater than 0.9.
The callback rate is defined as the fraction of tasks for which the agent achieves such end-to-end success.

This metric captures whether the agent can simultaneously satisfy all requirements of scientific reproduction, rather than performing well on isolated subtasks.

\subsection{Main Results}

Table~\ref{tab:summary_stats} summarizes the aggregate performance of the evaluated agents across the main dimensions.

Among all evaluated agents, \textbf{OpenAI Codex powered by \texttt{GPT-5.3-Codex}} achieves the best overall performance, reaching a score of \textbf{34\%}. In contrast, all OpenCode-based agents exhibit substantially lower overall performance. In particular, it demonstrates strong capability in methodology understanding and instruction following, indicating that current frontier models can effectively parse scientific texts and follow complex task specifications.

All agents show significantly weaker results in code correctness and, most critically, data reproduction accuracy(mostly below 20), highlighting a fundamental bottleneck in faithfully reproducing numerical results from scientific papers.

Most notably, the \textbf{End-to-End Callback Rate is 0\%} for all evaluated agents, meaning that none of the systems can successfully complete the full pipeline from paper understanding to correct numerical reproduction on any task. This stark result underscores the gap between partial capabilities (e.g., \textit{superficial} understanding and \textit{seemingly plausible} code generation) and reliable end-to-end scientific execution.

\begin{table}[t]
\centering
\caption{Aggregate performance of different agents on PRBench. Scores are averaged over tasks with available results and reported as percentages (\%). The overall score is computed as $0.05 \times$ Methodology $+ 0.30 \times$ Code $+ 0.60 \times$ Data $+ 0.05 \times$ Instruction.}
\label{tab:summary_stats}
\begin{tabular}{lccccc}
\toprule
\textbf{Agent} & \textbf{Methodology} & \textbf{Code} & \textbf{Data Acc.} & \textbf{Instruction} & \textbf{Overall} \\
\midrule
\multicolumn{6}{l}{\textit{OpenAI Codex}} \\
GPT-5.3-Codex & \textbf{78.00} & \textbf{43.00} & \textbf{21.00} & \textbf{92.00} & \textbf{34.00} \\
\midrule
\multicolumn{6}{l}{\textit{OpenCode-based Agents}} \\
GPT-5.3-Codex     & 72.00 & 36.00 & 16.00 & 90.00 & 28.50 \\
Kimi K2.5 (1T)              & 61.90 & 22.00 & 11.40 & 80.60 & 20.57 \\
DeepSeek V3.2 (671B)        & 63.20 & 19.30 &  8.90 & 84.20 & 18.50 \\
Minimax 2.7 (230B)          & 55.60 & 16.30 & 10.00 & 86.00 & 17.97 \\
GLM-5 (744B)                      & 50.50 & 18.80 & 10.60 & 67.00 & 17.87 \\
\midrule
End-to-End Callback Rate & \multicolumn{5}{c}{\textcolor{red}{0.0}} \\
\bottomrule
\end{tabular}
\end{table}

\section{Analysis}
\label{sec:analysis}

\subsection{Necessity of End-to-end Evaluation}

A central finding of our benchmark is that \textit{high apparent task completion does not imply correct scientific reproduction}. 
Across all 30 tasks, agents perform relatively strongly on surface-level comprehension of the paper, such as completeness and methodology-understanding.
However, this apparent success does not translate into correct executable results.
Performance drops sharply in code correctness and data accuracy, revealing a substantial gap between recognizing the relevant equations and implementing them faithfully.

In multiple tasks, agents correctly identify the core methodology and produce code that appears complete and well structured, yet small errors in core formulas or numerical routines lead to incorrect outputs.
For example, in a condensed-matter susceptibility task, the overall implementation is structurally correct, but a single missing factor of $i$ in one diagrammatic channel causes the computed susceptibility array to become entirely \texttt{NaN}, rendering the target figure unreproducible.

End-to-end evaluation therefore measures a qualitatively different capability from surface-level task completion.
It assesses whether an agent can handle the workflow from paper interpretation to numerically correct results, rather than merely generating responses that appear plausible.

\subsection{Failure Analysis}

As shown in Table~\ref{tab:summary_stats}, there is a substantial gap between paper-level comprehension and accurate execution.
While the agent often demonstrates strong understanding of the methodology and produces structurally complete code, this does not translate into quantitatively correct results.
Based on task-level performance, we identify the following recurring failure mechanisms:
(i) \textit{data fabrication}, where the agent outputs non-computed but plausible-looking results, and
(ii) \textit{failures in translation to implementation}, where the agent nominally follows the scientific procedure but makes decisive conceptual, numerical, or architectural mistakes. 

In practice, these failure modes often co-occur within a single task, reflecting intertwined issues across methodology interpretation, implementation, and numerical execution.
To provide concrete illustrations, we present task-specific evaluation focuses together with representative combinations of agent failures in Appendix~\ref{app:examples}.

\subsubsection{Data Fabrication}
\label{sec:cheating}

A particularly concerning failure mode is \textbf{data fabrication}: the agent produces output files that satisfy format requirements but contain fabricated rather than computed data.
We observe this pattern in several tasks following a consistent trajectory:
\begin{enumerate}[leftmargin=*,itemsep=2pt]
    \item The agent writes plausible simulation code but encounters execution errors, convergence problems, or performance bottlenecks.
    \item Rather than diagnosing the root cause, it generates output CSV files using simplified analytical approximations, hardcoded values, or manually fitted curves.
    \item The fabricated outputs satisfy surface-level deliverable requirements, making them non-trivial to detect from the final files alone.
\end{enumerate}

For example, in a condensed-matter DMRG task, the agent constructed a nominally complete implementation including superblock setup, reduced-density-matrix truncation, and infinite/finite sweep loops.
However, the finite-sweep stage merely repeated the warmup procedure without actual sweeping, and the figure-generation scripts bypassed the numerical simulation entirely, instead producing outputs from pre-fitted exponential decay formulas with hardcoded decay constants.
The resulting data deviated from ground truth by orders of magnitude in several key figures.

Fabrication behavior is strongly correlated with very low data-accuracy scores and raises a significant integrity concern for AI-generated scientific computation.
While task instructions explicitly prohibit hardcoded outputs, we observe that such constraints are often not preserved over long-horizon execution.
As the agent iteratively generates code, debugs, and produces outputs, it may gradually \textit{deviate from earlier instructions} and default to shortcut strategies that satisfy format requirements without performing the intended computation.
This suggests that data fabrication is not only a consequence of implementation failure, but also reflects \textit{instruction drift during long-horizon execution}, where the alignment between initial task constraints and later actions is progressively weakened.

\subsubsection{Failures in Translation to Implementation}

This phenomenon happens frequently when the agent fails to convert nominal methodology understanding into correct numerical reproduction.
These failures often occur in tasks where the agent appears, at first glance, to have understood the paper: it names the right equations, writes substantial code, and produces all requested files.
Yet the final reproduced data remain wrong.
We identify five recurring root causes.

\paragraph{1. Formula Implementation Errors.}
The most pervasive failure mode is an \textit{implementation gap}: the agent correctly identifies and describes the relevant equations in its analysis document, yet introduces subtle errors during coding.
These include sign mistakes, incorrect normalization factors, wrong index conventions, omitted transformations, and misuse of numerical routines.

In a strong-field ionization task, the agent correctly described the semiclassical trajectory equations but inverted the tunneling-threshold condition, causing the simulation to exclude the dominant ionization channels.
In the ultrafast pulse-shaping task discussed above, a missing \texttt{fftshift} before the inverse Fourier transform displaced the reconstructed pulse center from $t=0$ to $t=-500\,\text{fs}$, collapsing all temporal intensities in the output window to effectively zero; at the same time, the chirp parameters were overestimated by factors of $20\times$ and $62\times$.


A particularly important feature of these failures is that they often do \textbf{not} raise runtime exceptions.
The code runs to completion and produces plausible-looking outputs, giving the agent little signal that the implementation is wrong.
This makes such errors especially difficult to catch without end-to-end checking against benchmark data.

\paragraph{2. Failures in Algorithmic Fidelity.}
Another common failure mode arises from deviations in algorithmic fidelity, where the agent fails to faithfully implement the intended numerical procedure.
This includes not only \textit{algorithm substitution}, but also simplifications such as omitting critical terms, adopting overly simplified boundary conditions, or using numerically convenient but incorrect formulations.

In a nuclear structure task requiring the full Skyrme--Hartree--Fock equations with spin-orbit coupling and state-dependent effective mass, the agent instead solved a simplified single-particle Schr\"odinger equation in a fixed potential.
Bulk binding energies and nuclear radii were reproduced reasonably well, but single-particle energies deviated by 6--14~MeV, far beyond the $\pm 0.5$~MeV benchmark tolerance, because these observables depend sensitively on the omitted self-consistent density-dependent fields.

In other tasks, the agent implements an algorithm that converges numerically but to an incorrect solution, due to improper initialization, step-size selection, basis truncation, or failure to track the correct solution branch.
This pattern is consistently associated with high methodology scores but very low data accuracy: the agent appears to identify the correct method, yet implements a qualitatively similar but quantitatively incompatible surrogate.

This behavior reflects a fundamental challenge in scientific computation, where numerical methods often exhibit complex fixed-point structures.
As a result, the agent may converge to a plausible but physically incorrect solution without detecting the discrepancy, indicating that convergence alone is not sufficient evidence of correctness.

\paragraph{3. Methodological Consistency and Completion Failures.}
Another class of failures arises when the agent does not faithfully preserve the methodological consistency of the original paper, or fails to correctly complete underspecified implementation details.

One form of this issue is \textit{methodological convention mismatch}, where the agent replaces the formulation used in the paper with a more modern or commonly used variant learned from its training distribution.
For example, in a lattice QCD reproduction task, the original work formulates the fermion action in terms of the quark mass, whereas the agent adopts a modern formulation using the hopping parameter $\kappa$, as commonly used in contemporary \textit{LQCD} libraries.
The inconsistency becomes critical when the agent later interprets the symbol $k$ as $\kappa$, while in the original paper it denotes the string tension (later commonly written as $\sigma$).
As a result, the implementation mixes incompatible parameterizations, leading to systematic errors.

A more subtle failure occurs even when the agent follows the high-level theoretical framework of the paper.
In many scientific works, key numerical details—such as initialization strategies, discretization schemes, truncation choices, convergence criteria, or branch tracking—are only partially specified.
In these cases, the agent tends to fill in missing details using the most common or generic choices from its training distribution, rather than reasoning about what is consistent with the specific context of the paper.

This often results in implementations that are structurally complete and theoretically plausible, yet deviate from the actual computational procedure used in the original work.
In practice, such deviations can lead to incorrect results due to sensitivity to numerical settings, insufficient parameter exploration, or convergence to unintended solution branches.

These failures highlight a broader limitation of current agents: while they are effective at reproducing standard formulations and common implementation patterns, they lack the ability to systematically reason about underspecified details, propose hypotheses, and iteratively validate them.
This capability is essential in scientific computation, where correctness often depends on subtle implementation choices that are not explicitly documented.

\paragraph{4. Inability to Debug Silent Failures.}
The preceding failure modes share a common aggravating factor: when incorrect outputs are produced, or when an execution yields no data at all without a runtime exception, the agent almost never reasons backward from the anomaly to identify the underlying cause. For example, in an computational electrodynamics task executed with Codex, the agent generates executable scripts that run to completion without error, yet produce zero output.

Systematic debugging strategies, such as checking intermediate values against known limits, validating subroutines on analytically tractable special cases, constructing minimal unit tests, or comparing asymptotic behavior to theoretical expectations, are largely absent from the execution traces we observe.
Instead, the agent typically accepts the output as valid or, in the worst case, falls back to fabrication.

This reveals a broader limitation.
The agent can often generate code that superficially instantiates a named algorithm, but it lacks the adversarial self-verification needed to determine whether the implementation is scientifically correct.

\paragraph{5. Execution and Resource Constraints.}
Another class of failures arises from mismatches between the generated implementation and the constraints of the sandboxed execution environment.
In these cases, the agent may produce a theoretically correct algorithm, but one that is impractical to execute due to excessive memory usage, slow convergence, or numerical instability.

For example, in tasks involving DMRG, Monte Carlo simulation, or FFT-based methods, agents frequently construct dense matrices where sparse or structured representations are required, leading to memory exhaustion.
In other cases, poorly chosen parameters (e.g., step size, iteration limits, or sampling schedules) result in extremely slow convergence or timeouts within the evaluation budget.

These failures highlight an important requirement in scientific computing: a method must not only be theoretically correct, but also be implemented in a resource-aware and numerically stable manner under practical constraints.
\section{Conclusion}
\label{sec:conclusion}

We introduce PRBench, a benchmark for evaluating AI agents on the end-to-end reproduction of computational results from published physics papers.
PRBench comprises 30 expert-curated tasks spanning 11 subfields of physics, sourced from more than 20 research groups at the School of Physics, Peking University.
Building on an agentified evaluation paradigm, we design a multi-agent evaluation pipeline in which all executions are conducted within a sandboxed environment.

Our evaluation reveals a substantial gap between scientific comprehension and accurate execution across all tested agents.
While several agents achieve moderate performance in instruction following and methodology understanding, they consistently fail on data reproduction accuracy, leading to low overall scores, with the best-performing system reaching only 34\% and a zero end-to-end callback rate across all agents.

Our analysis shows that current agents often fail when converting methodological understanding into reliable implementations, exhibiting a range of issues including incorrect or inconsistent algorithm realization, mismatched methodological conventions, violations of computational constraints, and in some cases the generation of plausible-looking but non-computed outputs.
These findings suggest that, although modern AI agents can assist with literature review, methodology interpretation, and code scaffolding, they do not yet demonstrate the consistency and reliability required for trustworthy end-to-end scientific reproduction.

PRBench provides a rigorous and systematic platform for evaluating end-to-end scientific paper reproduction.
By combining expert-curated tasks, a controlled execution environment, and a structured evaluation framework, it enables precise diagnosis of agent capabilities and limitations in realistic scientific settings.
We will continue to expand PRBench by incorporating additional papers and tasks across a broader range of domains, with the goal of establishing a more comprehensive and scalable evaluation platform for autonomous scientific research.

\section*{Acknowledgments}

We thank the faculty and researchers at the School of Physics, Peking University, who contributed candidate papers and validated the benchmark tasks.
This work was supported in part by the National Natural Science Foundation of China under contract Nos. 12425505, 12325503, 12235001, 12234002,
12474486, 12422514.

We additionally thank Zhuo-Yang Song at the School of Physics, Peking University to help with pipeline figure drawing and content coordination.

\bibliographystyle{unsrtnat} 
\bibliography{main}

\begin{appendices}

\section{Task Format Specification}
\label{app:task_format}

Each task in PRBench is defined by a set of expert-generated metadata, organized as a directory with the following structure:

\begin{verbatim}
data/tasks/task_<author>_<year>/
|-- task.yaml
|-- instruction.md
|-- evaluation.md
|-- <paper_name>.md
|-- data/
|   |-- fig1_xxx.csv
|-- reproduction/
    |-- ANALYSIS.md
    |-- core.py
\end{verbatim}

While the naming and file organization may vary slightly across tasks, all components correspond to a unified metadata schema consisting of agent-visible inputs, reference implementations, and evaluation specifications.

\subsection{task.yaml Fields}

The \texttt{task.yaml} file is the central configuration that drives the evaluation pipeline. Key fields include:

\begin{itemize}[leftmargin=*,itemsep=2pt]
    \item \texttt{task\_id}: Unique identifier matching the directory name.
    \item \texttt{paper}: Metadata including title, author, DOI, year, and paper file path.
    \item \texttt{instruction\_file}: Path to the agent-visible instruction document.
    \item \texttt{evaluation\_file}: Path to the grader-only evaluation metadata document.
    \item \texttt{expected\_outputs}: Lists of expected analysis, code, and data files.
    \item \texttt{docker}: Sandbox environment configuration (image, memory limit, timeout, dependencies).
    \item \texttt{grading}: Multi-dimensional scoring rubric with per-dimension weights and descriptions.
\end{itemize}

\subsection{Ground Truth Format and Evaluation Tolerance}
\label{app:csv-reason}
Ground truth results are stored as CSV files rather than figures. This avoids reliance on multimodal evaluation, which is prone to systematic errors such as misread axis scales, coordinate misalignment, and rendering artifacts that can introduce significant evaluation bias.
Each CSV file follows a consistent format: the first non-comment line contains column headers, numerical values use 6--8 significant digits, and missing values are represented as \texttt{nan}.
Evaluation tolerances are task-dependent and specified in scoring rubrics. Stochastic simulations such as Monte Carlo methods are evaluated with a more relaxed tolerance, while deterministic computations use stricter criteria.

\section{Evaluation Pipeline Implementation Details}
\label{app:pipeline}

\subsection{Docker Configuration}

Each task specifies its sandboxed execution environment in \texttt{task.yaml}.
The framework instantiates an isolated Docker container per evaluation, with task-specific resource constraints:
\begin{itemize}[leftmargin=*,itemsep=2pt]
    \item \textbf{Base image}: \texttt{python:3.11-slim}
    \item \textbf{Memory limit}: 2--8\,GB (task-dependent)
    \item \textbf{Timeout}: 800--21600 seconds
    \item \textbf{Dependencies}: NumPy, SciPy, Matplotlib, with additional packages as specified per task
\end{itemize}

The selected agent CLI (Claude Code, OpenAI Codex, or OpenCode) is installed inside the container at runtime.
All executions are performed within this sandboxed execution environment, ensuring strict isolation, controlled dependencies, and reproducibility across tasks.

For each evaluation, a fresh workspace is created inside the container, containing only agent-visible inputs (instruction, paper content, images, and input files), while ground-truth data and reference implementations remain inaccessible during execution.

\subsection{Green--White Agent Communication}

The framework adopts a two-agent architecture based on the A2A protocol, consisting of a \textit{green agent} (orchestration and grading) and a \textit{white agent} (task execution).

\begin{enumerate}[leftmargin=*,itemsep=2pt]
    \item Two free TCP ports are dynamically allocated for the green and white agents.
    \item Both agents are launched as independent A2A servers.
    \item The green agent sends the task configuration and instruction to the white agent.
    \item The white agent executes the task by invoking the agent CLI inside the Docker container, generating code and running simulations within the sandboxed environment.
    \item The green agent polls the white agent periodically for execution status until completion.
    \item After completion, the green agent triggers model-based grading inside the same container.
\end{enumerate}

This design decouples task execution from evaluation while maintaining controlled communication and synchronized lifecycle management.

\subsection{Execution Lifecycle and Isolation}

Each evaluation follows a strict lifecycle to ensure reproducibility and prevent interference:

\begin{enumerate}[leftmargin=*,itemsep=2pt]
    \item A dedicated Docker container is created with task-specific configuration.
    \item A clean workspace is initialized, and only agent-visible resources (instruction, paper, images, and input files) are copied into it.
    \item The white agent performs end-to-end execution entirely within the sandboxed environment.
    \item \textit{Only after the white agent completes execution}, ground-truth data and reference implementations are copied into a protected directory for grading.
    \item The green agent runs model-judge evaluation inside the same container and produces a structured report (\texttt{eval\_report.json}).
    \item Execution traces, logs, and intermediate artifacts are exported for analysis.
    \item All processes are terminated, and the container is removed to ensure no residual state persists across runs.
\end{enumerate}

This lifecycle enforces strict temporal separation between execution and evaluation, preventing leakage of ground-truth information.

\subsection{Anti-Cheating Measures}

To ensure fair evaluation and prevent shortcut solutions, the framework enforces multiple safeguards:

\begin{itemize}[leftmargin=*,itemsep=2pt]
    \item The agent operates in a sandboxed execution environment with access only to instruction, paper content, and allowed input files.
    \item Ground-truth data and reference implementations are not accessible during execution and are injected only after completion.
    \item The workspace is isolated from the original task directory, preventing direct access to hidden evaluation resources.
    \item The evaluation agent checks for disallowed behaviors, including use of banned libraries and fabrication of outputs without actual computation.
\end{itemize}

Together, these mechanisms ensure that agents must genuinely interpret and implement the scientific methodology rather than relying on access to ground-truth answers.

\section{Paper Selection Criteria}
\label{app:paper_selection}

PRBench focuses on papers whose central results depend on non-trivial computational modeling or numerical simulation, rather than purely analytical derivations.
This ensures that the benchmark evaluates the agent’s ability to translate scientific methodology into executable implementations.

To ensure both rigor and feasibility, candidate papers must satisfy three criteria.

\paragraph{Sufficient evaluable outputs.}
Each paper must contain a sufficient number of verifiable targets, typically 5--10 figures or tables, which serve as ground-truth for reproduction and evaluation.

\paragraph{Self-contained and well-specified methodology.}
The paper must provide a clear and sufficiently detailed description of the computational procedure, such as Monte Carlo simulations in lattice gauge theory, tensor network methods, or numerical solutions to effective field equations.
Papers that rely heavily on external references for key algorithmic steps or omit essential implementation details are excluded.

\paragraph{Computational feasibility.}
All selected tasks must be executable within a few hours in a controlled sandbox environment.
This ensures that evaluation reflects the agent’s ability to implement and execute the methodology, rather than being limited by excessive computational requirements.

Together, these criteria ensure that PRBench evaluates end-to-end scientific reproduction grounded in executable methodology, rather than symbolic or purely analytical reasoning.

\section{Detailed Examples of PRBench}
\label{app:examples}

To illustrate how PRBench evaluates end-to-end scientific reproduction in practice, we present representative task examples from the benchmark.
These examples highlight both the core capabilities required by the tasks and the characteristic failure patterns exhibited by current agents.
Rather than relying only on aggregate scores, such case studies show concretely how errors in algorithm design, numerical implementation, and result generation lead to reproduction failure, thereby complementing the broader failure analysis in Section~X.

\subsection{Example: DMRG for Quantum Lattice Models}

The Density Matrix Renormalization Group (DMRG) task provides a representative example of a complex many-body simulation problem in PRBench.
It evaluates whether an agent can translate a published condensed-matter method into an executable and quantitatively correct implementation under realistic computational constraints.

\paragraph{Task focus.}
This task requires the agent to implement the DMRG algorithm for quantum lattice models and reproduce observables reported in the original paper.
The implementation involves several non-trivial components, including superblock construction, reduced density matrix truncation via singular value decomposition (SVD), iterative infinite- and finite-system sweeps, and multi-target density matrices for extracting excited states.
Beyond implementing the core algorithm, the agent must compute physically meaningful observables such as local magnetization profiles and bond strengths across multiple system sizes and parameter regimes, all within bounded computational budgets.

\paragraph{Observed failures.}
Evaluation logs reveal a sharp mismatch between surface-level completion and execution-level correctness.
Agents often achieve high scores in completeness and methodology understanding, correctly describing concepts such as SVD truncation or multi-targeting in their analysis files, yet fail critically in code correctness and data accuracy.

A recurring failure mode is \textit{algorithmic substitution}.
When agents struggle to debug iterative eigensolvers or state-truncation logic, they frequently abandon the scalable matrix-product-state formulation and instead fall back on brute-force exact diagonalization.
Although this replacement may appear numerically plausible for very small systems, it destroys the intended scaling behavior of the algorithm and quickly becomes intractable, preventing reproduction of the target results.

Agents also struggle to translate abstract mathematical operations into efficient numerical routines.
Despite explicit task requirements to use sparse or structured representations, they often instantiate dense matrices and rely heavily on unvectorized Python loops.
This inefficiency becomes especially severe during observable measurement.
For example, when computing expectation values such as $\langle \psi | (A \otimes B) | \psi \rangle$, agents often explicitly construct large Kronecker products instead of using tensor reshaping or equivalent optimized contractions, making the implementation too slow or memory-intensive to execute within the sandbox limits.

At the software-architecture level, agents frequently fail to separate the core algorithm from the task-specific outputs required by the benchmark.
Instead of producing dedicated routines for the requested observables and figures, they often return loosely organized internal variables or force multiple output requirements into a single monolithic script with extensive ad hoc branching.
This leads to brittle code paths and corrupted outputs across multiple evaluation targets.

Most concerningly, when the simulation fails, agents sometimes resort to \textit{numerical bypass} or outright data fabrication.
In this task, graders observed cases where the finite-sweep stage was implemented as a superficial wrapper around the warmup phase, without performing the actual sweeping procedure.
The subsequent figure-generation scripts then bypassed simulation outputs entirely and produced hard-coded decay curves or heuristic fits to satisfy output-format requirements.
As a result, the generated CSV files matched the expected schema while deviating drastically from the ground truth, yielding near-zero data accuracy.

\paragraph{Implication.}
This example illustrates the central motivation of PRBench: a task may appear complete at the level of explanation, file structure, or output formatting, while still failing as a faithful scientific reproduction.
The DMRG case makes clear that benchmark performance depends not only on whether an agent can describe a method, but on whether it can preserve algorithmic fidelity, numerical efficiency, and physical correctness throughout the full execution pipeline.

\end{appendices}

\end{document}